\documentclass[11pt]{article}
\usepackage[preprint]{acl}
\usepackage{conference_ACL}
 \usepackage[table]{xcolor}
 \usepackage{booktabs}
 \usepackage{multirow}
 \usepackage{nicefrac}

 \newcommand{\qcol}{\cellcolor{blue!5}}
 
\title{More Than Bits: Multi-Envelope Double Binary Factorization \\
for Extreme Quantization}

\author{
 \textbf{Yuma Ichikawa\textsuperscript{1, 2}},
 \textbf{Yoshihiko Fujisawa\textsuperscript{1, 3}},
 \textbf{Yudai Fujimoto\textsuperscript{1, 3}},
 \\
 \textbf{Akira Sakai\textsuperscript{1, 4}},
 \textbf{Katsuki Fujisawa\textsuperscript{3}}
 \\
 \\
 \textsuperscript{1}Fujitsu Limited,
 \textsuperscript{2}RIKEN Center for AIP,
 \textsuperscript{3}Institute of Science Tokyo,
 \textsuperscript{4}Tokai University
 \\
 \small{
   \textbf{Correspondence:} \href{mailto:ichikawa.yuma@fujitsu.com}{ichikawa.yuma@fujitsu.com}
 }
}

\begin{document}
\maketitle

\begin{abstract}
    For extreme low-bit quantization of large language models (LLMs), Double Binary Factorization (DBF) is attractive as it enables efficient inference without sacrificing accuracy.
    However, the scaling parameters of DBF are too restrictive; after factoring out signs, all rank components share the same magnitude profile, resulting in performance saturation.
    We propose \textbf{Multi-Envelope DBF (MDBF)}, which retains a \emph{shared} pair of 1-bit sign bases but replaces the single envelope with a rank-$l$ envelope.
    By sharing sign matrices among envelope components, MDBF effectively maintains a binary carrier and utilizes the limited memory budget for magnitude expressiveness.
    We also introduce a closed-form initialization and an alternating refinement method to optimize MDBF.
    Across the LLaMA and Qwen families, MDBF enhances perplexity and zero-shot accuracy over previous binary formats at matched bits per weight while preserving the same deployment-friendly inference primitive.
\end{abstract}

\section{Introduction}\label{sec:intro}
Large language models (LLMs) support many NLP systems; however, their size renders deployment expensive, since storing FP16 or FP32 parameters and moving them through the memory hierarchy often dominates both the memory footprint and inference latency.
Quantization is therefore a central tool for efficient deployment. 
Post-training quantization (PTQ) \citep{frantar2022gptq, lin2024awq} is particularly appealing because it can be applied to a pretrained model with minimal overhead, avoiding the need for full retraining. 
Although recent PTQ methods maintain strong accuracy at around 4-bit precision, performance typically degrades as precision approaches the 2--1-bit regime, where the per-layer information budget is extremely limited. 
To push below 2 bits lower, many approaches move beyond elementwise quantization and adopt structured parameterizations \citep{chee2023quip, tseng2024qtip, tseng2024quip, malinovskii2024pvtuning}. 
Binary and near-binary schemes are especially appealing because they provide a clear hardware fast path: most computation can be performed by specialized kernels operating on bit-packed sign matrices, with only lightweight higher-precision scaling.

A prominent family of methods factorizes each weight matrix into low-rank components and then binarizes the factors. 
OneBit \citep{xu2024onebit} shows that appropriate scaling can stabilize 1-bit factors, while Double Binary Factorization (DBF) \citep{boza2025dbf} makes the binary path explicit by composing two binary matrix multiplications with interleaved diagonal scalings. 
LittleBit \citep{lee2025littlebit} further enhances extreme-bit accuracy through multi-scale scaling and residual compensation, utilizing quantization-aware training (QAT) across multiple GPUs. 
Despite these advances, existing formats share a key structural limitation: after demodulation, factor magnitudes are confined to a single rank-one envelope.
Increasing the inner rank primarily enhances sign diversity rather than magnitude expressiveness.
As a result, under a fixed bits-per-weight budget, accuracy can saturate because gains come more from signs than magnitudes.

\begin{figure*}[tb]
    \centering
    \includegraphics[width=\linewidth]{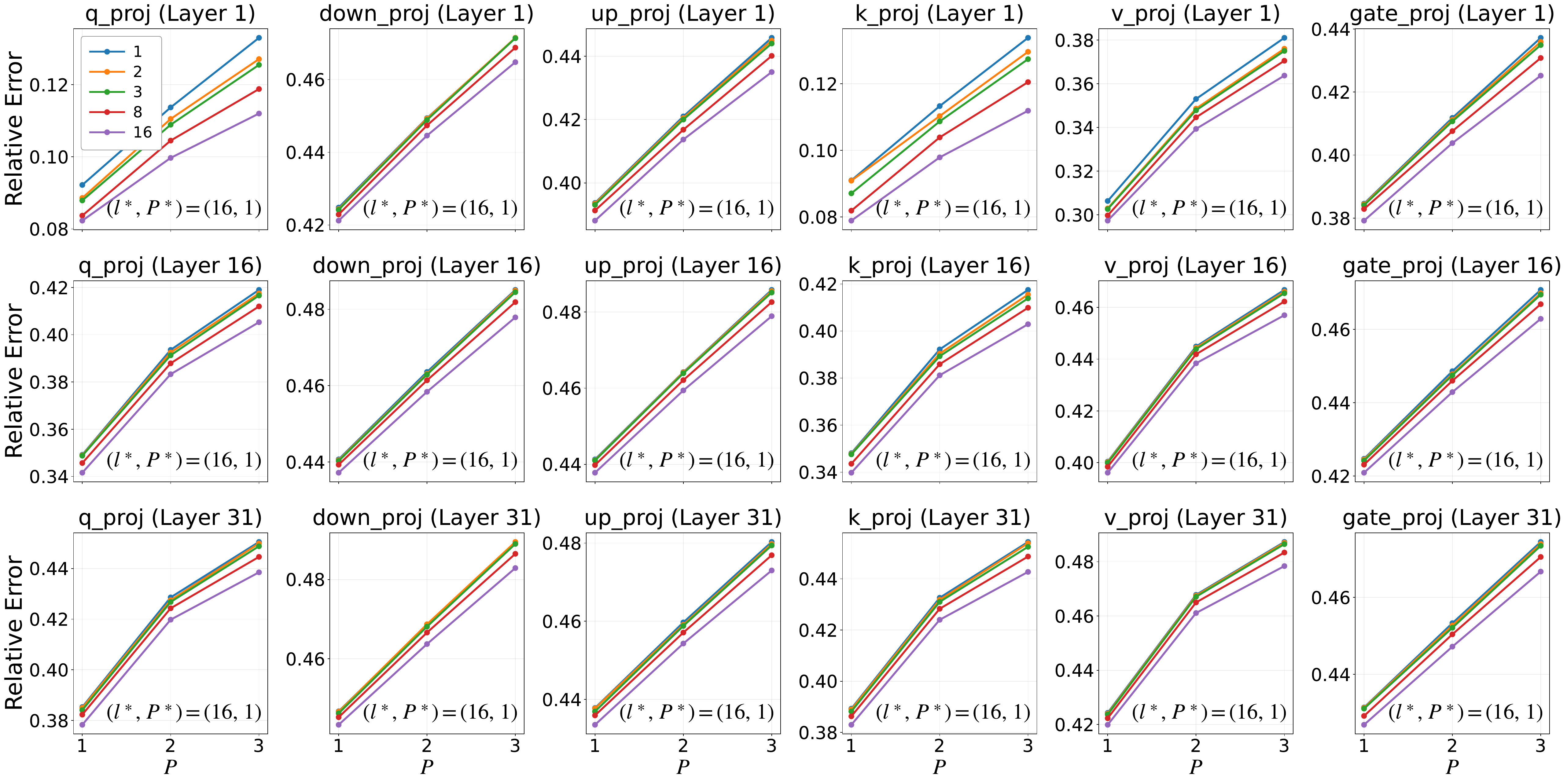}
    \caption{\textbf{Layer-wise reconstruction error vs.\ envelope rank and decomposition depth (LLaMA2 7B).}
    We conduct experiments under a 1.5 bit quantization setting and report the relative Frobenius error $\nicefrac{\|W-\widehat{W}\|_{F}}{\|W\|_{F}}$ of MDBF as a function of the envelope rank $l$ and the decomposition depth $P$, evaluated on three representative Transformer blocks of layers $0$, $15$, and $30$ and central attention/MLP projections.
    Across layers and modules, larger $l$, which increases magnitude expressivity while sharing sign bases, consistently lowers reconstruction error. In contrast, larger $P$, which adds extra decompositions and sign bases, often worsens reconstruction at matched bits per weight. The best configuration in this sweep is $(l^\ast, P^\ast)=(16,1)$.}
    \label{fig:relerr_lP_llama7b}
\end{figure*}

This paper addresses the DBF bottleneck by explicitly allocating limited expressivity to the critical components at extremely low precision.
We propose \emph{Multi-Envelope Double Binary Factorization} (MDBF), which retains the shared 1-bit sign bases and a deployment-friendly binary fast path while replacing the rank-one magnitude envelope with multiple demodulated envelope modes.
Figure~\ref{fig:relerr_lP_llama7b} shows that increasing the envelope rank $l$ systematically reduces reconstruction error; however, while increasing the residual path $P$, as in LittleBit \citep{lee2025littlebit}, is often less effective within a fixed bits-per-weight budget.
MDBF adds a small number of real-valued degrees of freedom for magnitude modeling, better aligning with the empirically observed low-rank structure of Transformer weights, which are rarely rank-one.
To make MDBF applicable for layer-wise PTQ, we introduce a layer-wise optimization pipeline with closed-form initialization followed by ADMM refinement.
Across the LLaMA and Qwen families, MDBF improves perplexity and zero-shot accuracy compared to previous binary formats at matched bits per weight, particularly in the challenging 2–1 bit range, while maintaining the same deployment-friendly binary inference primitive.

\paragraph{Contributions.}
\begin{itemize}
    \item \textbf{Identifying DBF Bottleneck:} We identify the bottleneck as the \emph{single envelope constraint}. Under a fixed bits-per-weight budget, modeling \emph{magnitude} variation yields greater accuracy gains than increasing \emph{sign} diversity.
    \item \textbf{Multi-Envelope Generalization of DBF:} We propose Multi-Envelope DBF (MDBF), which retains the shared 1-bit sign bases and maintains the same deployment-friendly binary fast path while replacing the rank-one magnitude envelope with a rank-$l$ envelope.
    \item \textbf{Initialization and ADMM Refinement for MDBF:} We generalize the initialization of LittleBit and the ADMM-based refinement of DBF to a multi-envelope setting. This results in a closed-form initializer and an efficient alternating ADMM refinement procedure.
    \item \textbf{Empirical Validation:} Across the LLaMA and Qwen model families, MDBF consistently reduces reconstruction error and improves perplexity and zero-shot accuracy compared to prior binary formats at matched BPW.
\end{itemize}

\section{Notation}
Vectors are denoted by bold lowercase letters, e.g., $\B{x}$, and matrices by uppercase letters, e.g., $W$. Throughout, $W\in\mab{R}^{N\times M}$ denotes a real-valued weight matrix. We denote by $\odot$ the Hadamard elementwise product, and for a vector $\B{a}$, let $D_{\B{a}}$ represent the diagonal matrix with $(D_{\B{a}})_{ii}=a_i$. We use $\|\cdot\|_F$ for the Frobenius norm and $\langle A,B\rangle_{F}\coloneqq \mathrm{Tr}(A^\top B)$ for the corresponding Frobenius inner product.
For any matrix $A$, its singular values are denoted by $\sigma_1(A)\ge \cdots \ge \sigma_{\min(N,M)}(A)\ge 0$. For a target rank $R\le \min(N,M)$, we denote the rank-$R$ truncated SVD of $W$ as $W_R = U_R\Sigma_R V_R^\top$, where $U_R\in\mab{R}^{N\times R}$ and $V_R\in\mab{R}^{M\times R}$ have orthonormal columns and $\Sigma_R=\mathrm{diag}(\sigma_1,\dots,\sigma_R)\in\mab{R}^{R\times R}$. Finally, the entrywise sign function $\mathrm{sign}(\cdot)$ maps to $\{\pm 1\}$, with $\mathrm{sign}(0)=+1$.

\section{Preliminaries}
\label{sec:Preliminaries}

\subsection{Low-Rank Approximation}
\label{subsec:low-rank-approximation}
A common approach to model compression exploits the empirically observed approximate low-rank structure of weight matrices. 
Given a weight matrix $W \in \mab{R}^{N\times M}$, we approximate it with a rank-$R$ factorization:
\begin{equation}
    \label{eq:low-rank-basis}
    W \approx U V^{\top},~~
    U \in \mab{R}^{N \times R},~V \in \mab{R}^{M \times R}.
\end{equation}
Equivalently, $UV^{\top}$ expresses $W$ as a sum of $R$ rank-one components,
$UV^{\top}=\sum_{j=1}^{R}\B{u}_{j}\B{v}_{j}^{\top}$, where $\B{u}_{j}$ and $\B{v}_{j}$ are the $j$-th columns of $U$ and $V$.

However, low rank alone does not guarantee meaningful memory savings. 
If $U$ and $V$ are stored in standard high-precision formats, such as FP16 or FP32, the parameter count becomes $(N+M)R$, which may be comparable to the original $NM$ parameters unless $R \ll \min(N, M)$. 
Moreover, maintaining accuracy often requires a moderate $R$, further limiting the compression benefit.
Therefore, to achieve significant savings at very low bit widths while maintaining a sufficiently large effective rank, it is crucial to quantize the factors or introduce additional structures that facilitate both compact storage and efficient computation.

\subsection{Double Binary Factorization}
Double Binary Factorization (DBF)~\citep{boza2025dbf} represents a weight matrix $W\in\mab{R}^{N\times M}$ by utilizing two binary sign bases and diagonal rescalings:
\begin{equation}
    \label{eq:dbf-basis}
    \widehat{W}_{\mathrm{DBF}} \coloneqq D_{\B{a}} S_{a} D_{\B{m}} S_{b}^{\top} D_{\B{b}},
\end{equation}
where $\B{a}\in\mab{R}^{N}$ and $\B{b}\in\mab{R}^{M}$ are the row and column scaling vectors,
$\B{m}\in\mab{R}^{R}$ is an inner scaling vector, and $S_{a}\in\{\pm 1\}^{N\times R}$ and $S_{b}\in\{\pm 1\}^{M\times R}$ are binary sign matrices.
Equation~\eqref{eq:dbf-basis} can be regarded as a structured rank-$R$ factorization:
\begin{multline}
    \label{eq:dbf-factors}
    \widehat{W}_{\mathrm{DBF}}=\widehat{U}\widehat{V}^{\top},\\
    \widehat{U} \coloneqq D_{\B{a}} S_{a} D_{\B{m}^{\nicefrac{1}{2}}},~
    \widehat{V} \coloneqq D_{\B{b}} S_{b} D_{\B{m}^{\nicefrac{1}{2}}}.    
\end{multline}

\paragraph{Single-Envelope Constraint.}
For a sign mask $S \in \{\pm1\}^{p\times q}$ and a matrix $Z\in\mab{R}^{p\times q}$,
we define the \emph{demodulated envelope} of $Z$ concerning $S$ as $E_S(Z) \coloneqq S\odot Z$.
In DBF, the demodulated envelopes of factors $\widehat{U}$ and $\widehat{V}$ satisfy the following exact identities:
\begin{multline}
    \label{eq:dbf-demod-envelope}
    E_{S_a}(\widehat{U}) = S_a\odot\widehat{U} = \B{a}(\B{m}^{\nicefrac{1}{2}})^{\top}, \\
    E_{S_b}(\widehat{V}) = S_b\odot\widehat{V} = \B{b}(\B{m}^{\nicefrac{1}{2}})^{\top}.
\end{multline}
Therefore, the following holds:
\begin{equation}
    \mathrm{rank}(E_{S_a}(\widehat{U}))\le 1,~~
    \mathrm{rank}(E_{S_b}(\widehat{V}))\le 1,
\end{equation}
after demodulation by the shared sign bases, both factor envelopes constitute rank-one outer products.
Equivalently, each column of $\widehat{U}$ shares the same row-wise envelope profile $\B{a}$
up to a scalar multiplier $m_j^{1/2}$, and similarly $\B{b}$ applies to $\widehat{V}$.
We refer to this rank-one demodulated-envelope constraint as the \emph{single envelope constraint}.

\paragraph{Inference.}
Let $X\in\mab{R}^{T\times N}$ denote the input activations and
$Y=X\widehat{W}_{\mathrm{DBF}}\in\mab{R}^{T\times M}$ the output.
DBF admits an efficient evaluation order in which diagonal scalings are interleaved with
two binary matrix multiplications:
\begin{equation}
    \label{eq:DBF-inference-form}
    Y = X\widehat{W}_{\mathrm{DBF}}
    = \Bigl(\Bigl(\Bigl(\bigl(X D_{\B{a}}\bigr) S_{a}\Bigr) D_{\B{m}}\Bigr) S_{b}^{\top}\Bigr) D_{\B{b}}.
\end{equation}
Multiplication by a diagonal matrix is equivalent to elementwise rescaling and is typically bandwidth-light compared to matrix multiplication. 
Therefore, DBF substitutes one high-precision GEMM with two binary GEMMs and inexpensive diagonal operations. 
The effective bits-per-weight is determined by the storage of bitpacked $\{S_a,S_b\}$ and real-valued vectors $\{\B{a},\B{b},\B{m}\}$, relative to the $NM$ entries of the original matrix.

\paragraph{Residual Compensation.}
LittleBit \citep{lee2025littlebit} improves fidelity by augmenting a single DBF approximation with an additional DBF term trained
on the residual. 
LittleBit specifically employs a two-term decomposition as follows:
\begin{multline}
\label{eq:dbf-residual-comp}
    \widehat{W}_{\mathrm{LittleBit}}
    =D_{\B{a}^{(1)}} S_{a}^{(1)} D_{\B{m}^{(1)}} S_{b}^{(1) \top} D_{\B{b}^{(1)}}
    \\
    +D_{\B{a}^{(2)}} S_{a}^{(2)} D_{\B{m}^{(2)}} S_{b}^{(2) \top} D_{\B{b}^{(2)}}.
\end{multline}
A natural optimization strategy is stagewise residual fitting: first, fit the initial term, then compute the residual term:
\begin{equation}
    \widehat{R} \coloneqq W-D_{\B{a}^{(1)}} S_{a}^{(1)} D_{\B{m}^{(1)}} S_{b}^{(1) \top} D_{\B{b}^{(1)}}.
\end{equation}
Fit the second DBF term to approximate $\widehat{R}$.
This residual-compensation scheme captures structures that are difficult to represent with a single DBF component.

\section{Method}\label{sec:method}

\begin{figure}[tb]
    \centering
    \includegraphics[width=\linewidth]{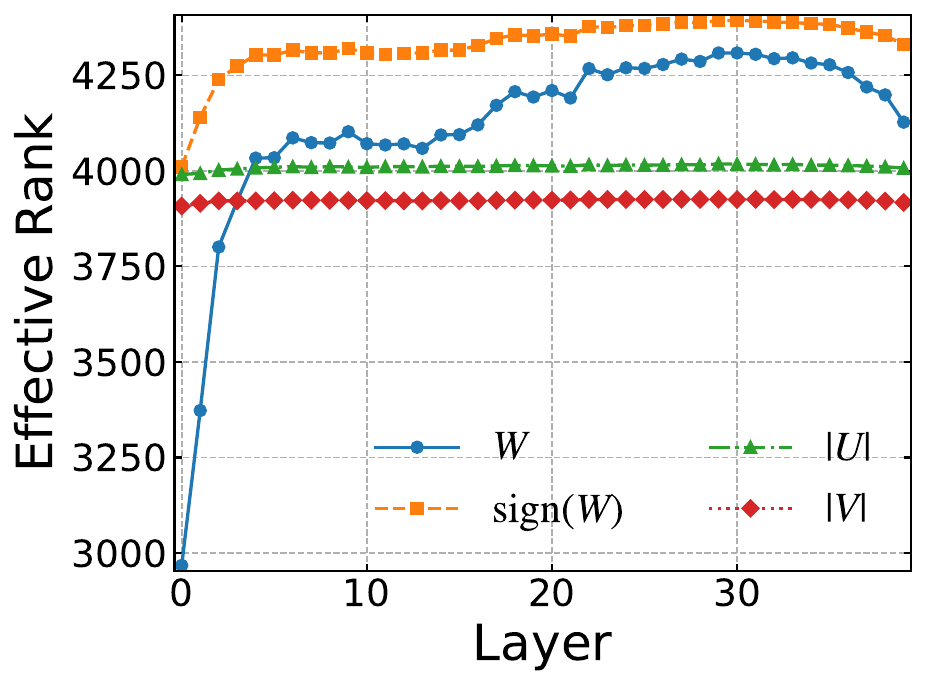}
    \caption{Entropy-based effective rank across Transformer layers for LLaMA2 13B. We report the average effective rank of the full weight matrices $W$, their binarized signs $\mathrm{sign}(W)$, and the demodulated envelopes $|U|$ and $|V|$. The envelopes remain consistently above rank one, indicating multiple magnitude modes and motivating the relaxation of the single-envelope constraint.}
    \label{fig:effective-rank}
\end{figure}

\subsection{Bottleneck: Single-Envelope Constraint}
\label{subsec:bottleneck_single_envelope}

In practice, DBF exhibits a performance ceiling; increasing the inner dimension $R$ results in diminishing returns, and accuracy saturates below that of a real-valued low-rank approximation with a comparable parameter count. 
This gap is not merely empirical but structural. 
As shown in Section \ref{sec:Preliminaries}, demodulation indicates that each DBF factor is constrained by a rank-one amplitude envelope shared across all $R$ columns. 
Enlarging $R$ primarily increases sign-pattern diversity rather than enhancing the capacity to represent magnitudes, fundamentally limiting achievable accuracy.

The assumption of a rank-one envelope is inconsistent with the empirical structure of LLM weights. To quantify this point, we analyze LLaMA2 13B and compute the entropy-based effective ranks \citep{roy2007effectiveRank} of the demodulated envelopes $|U|$ and $|V|$; see Appendix~\ref{app:effective_rank} for a detailed definition. Figure~\ref{fig:effective-rank} shows the average effective rank across Transformer blocks and their linear layers. Although the envelopes are low-rank, they are \emph{consistently not rank-one}; their effective ranks remain moderately above 1, suggesting multiple magnitude modes that DBF cannot represent under the single-envelope constraint.

These observations clarify why merely increasing sign variation, such as in residual-style multi-term designs like LittleBit, is not the most direct solution. The goal is to preserve a deployment-friendly inference path, where shared binary carriers facilitate efficient 1-bit computation while alleviating the bottleneck. This motivation leads to \emph{Multi-Envelope DBF}, which enhances expressivity by allowing higher-rank amplitude envelopes to be used under the same shared binary sign bases, without altering the inference-time execution model.

\begin{takeawaybox}
    \textbf{More Than Bits:} \emph{DBF is fundamentally limited by its rank-one amplitude envelope, rather than by the available bit budget.}
\end{takeawaybox}

\subsection{Multi-Envelope Double Binary Factorization}
We propose Multi-Envelope DBF (MDBF), which preserves the \emph{shared 1-bit sign bases} while allowing the demodulated envelopes of the factor matrices to have a rank of at most $l$ as follows:
\begin{equation}
    \label{eq:me-dbf-basis}
    \widehat{W}_{l}\coloneqq \widehat{U}_{l}\widehat{V}_{l}^{\top},
    \widehat{U}_{l} = S_{a}\odot A Q^{\top},
    \widehat{V}_{l} = S_{b}\odot B G^{\top},
\end{equation}
where $S_{a}\in\{\pm 1\}^{N\times R}$ and $S_{b}\in\{\pm 1\}^{M\times R}$ are binary sign matrices, while
$A\in\mab{R}^{N\times l}$, $Q\in\mab{R}^{R\times l}$, $B\in\mab{R}^{M\times l}$, and $G\in\mab{R}^{R\times l}$ are real-valued factors.
By construction, the demodulated envelopes are low-rank:
\begin{multline}
    \label{eq:me-dbf-envelope-rank}
    E_{S_a}(\widehat{U}_l)=S_a\odot \widehat{U}_l = AQ^{\top}, \\
    E_{S_b}(\widehat{V}_l)=S_b\odot \widehat{V}_l = BG^{\top},
\end{multline}
leads to $\mathrm{rank}(E_{S_a}(\widehat{U}_l))\le l$ and $\mathrm{rank}(E_{S_b}(\widehat{V}_l))\le l$.
Let $A=[\B{a}^{(1)},\dots,\B{a}^{(l)}]$ and $Q=[\B{q}^{(1)},\dots,\B{q}^{(l)}]$, along with
 $B=[\B{b}^{(1)},\dots,\B{b}^{(l)}]$ and $G=[\B{g}^{(1)},\dots,\B{g}^{(l)}]$, yield the expansions
\begin{equation}
\label{eq:me-dbf-expand}
    \widehat{U}_{l}=\sum_{t=1}^{l} D_{\B{a}^{(t)}} S_{a} D_{\B{q}^{(t)}},
    \widehat{V}_{l}=\sum_{s=1}^{l} D_{\B{b}^{(s)}} S_{b} D_{\B{g}^{(s)}}.
\end{equation}
Thus, $\widehat{W}_{l}$ admits an equivalent sum-of-DBF form:
\begin{equation}
\label{eq:me-dbf-sum}
    \widehat{W}_{l}
    =
    \sum_{t=1}^{l}\sum_{s=1}^{l}
    D_{\B{a}^{(t)}} S_{a} D_{\B{q}^{(t)} \odot \B{g}^{(s)}} S_{b}^{\top} D_{\B{b}^{(s)}}.
\end{equation}
MDBF generalizes DBF: when $l=1$, Equation~\eqref{eq:me-dbf-basis} reduces to a single DBF term involving
$\B{m}=\B{q}^{(1)}\odot \B{g}^{(1)}$; if one enforces $\B{m}\ge 0$, 
$\B{q}^{(1)}=\B{g}^{(1)}=\B{m}^{\nicefrac{1}{2}}$ may be set to column-sign absorption in $S_{a},S_{b}$.

\subsection{Multi-Envelope Structure}
\label{sec:multi-envelope-structure}

We next formalize the role of envelope rank $l$ and derive the corresponding closed-form optimizer using a fixed sign mask.
The key idea is to decompose each factor into (i) a \emph{binary sign mask} and (ii) a \emph{demodulated envelope}, while imposing a rank constraint on the envelope.

\paragraph{Demodulation under Fixed Sign Mask.}
Fix a sign mask $S\in\{\pm1\}^{N\times R}$ and consider an arbitrary factor matrix $U\in\mab{R}^{N\times R}$.
We define the \emph{demodulated envelope} as follows:
\begin{equation}
    E_{S}(U) \coloneqq S \odot U.
\end{equation}
Since $S_{ij}^{2}=1$ holds for all $(i,j)$, we obtain $S\odot(S\odot Z)=Z$ for any matrix $Z$.
Thus, once $S$ is fixed, $U$ is determined by its envelope $E_S(U)$.

Consider an approximation $\widehat{U}$ that \emph{shares the same mask} $S$. It can be expressed as $\widehat{U}=S\odot \widehat{E}$ for some envelope matrix $\widehat{E}\in\mab{R}^{N\times R}$.
Moreover, by Lemma~\ref{lem:sign-invariance}, the mapping $Z\mapsto S\odot Z$ preserves Frobenius norms:
\begin{equation}
    \label{eq:envelope-isometry}
    \|U-\widehat{U}\|_F =
    \|E_{S}(U)-\widehat{E}\|_F.
\end{equation}
Therefore, under a fixed mask $S$, approximating $U$ in Frobenius norm is equivalent to approximating its demodulated envelope.

\paragraph{Rank-$l$ Envelope Class.}
MDBF constrains the envelope to a low rank. For $l\ge 1$, we define the feasible family as follows:
\begin{definition}
    \label{def:Fl}
    Let $S\in\{\pm1\}^{N\times R}$ and $l\ge 1$.
    Define
    \begin{equation}
    \mac{F}_l(S)\coloneqq
    \ab\{S\odot E  \mid E\in\mab{R}^{N\times R}, \mathrm{rank}(E) \le l \}.
    \end{equation}
\end{definition}
Equivalently, $\widehat{U}\in\mac{F}_l(S)$ if and only if its envelope $E_S(\widehat{U})=S\odot \widehat{U}$ has a rank of at most $l$.
This implies the MDBF factor structure; for example, $\widehat{U}_l=S_a\odot(AQ^\top)$ satisfies $\mathrm{rank}(E_{S_a}(\widehat{U}_l))=\mathrm{rank}(AQ^\top)\le l$.

\paragraph{Closed-form Minimizer.}
By Equation~\eqref{eq:envelope-isometry}, the best approximation of $U$ within $\mac{F}_l(S)$ becomes the classical best rank-$l$ approximation problem for the envelope matrix $E_S(U)$. The following theorem demonstrates this equivalence and provides both the optimal error and a minimizer.

\begin{theorem}
    \label{thm:rankl-error-main}
    Let $U\in\mab{R}^{N \times R}$ fix a sign mask $S\in\{\pm 1\}^{N \times R}$.
    Let $\sigma_1(E_S(U))\ge \cdots \ge \sigma_{\min(N,R)}(E_S(U))\ge 0$
    be the singular values of $E_S(U)=S\odot U$.
    Fix an integer $l$ such that $1\le l\le \min(N, R)$.
    Then
    \begin{equation}
    \label{eq:rankl-error}
    \min_{\widehat{U}\in\mac{F}_l(S)} \|U-\widehat{U}\|_{F}
    =
    \ab(\sum_{i=l+1}^{\min(N, R)}\sigma_i\bigl(E_S(U)\bigr)^2)^{\nicefrac{1}{2}}.
    \end{equation}
    Moreover, one minimizer is
    \begin{equation}
        \label{eq:Ustar}
        \widehat{U}^{\ast}=S\odot \mathrm{TSVD}_{l}(E_{S}(U)),
    \end{equation}
    where $\mathrm{TSVD}_l(\cdot)$ denotes the rank-$l$ truncated SVD, which provides the best approximation in the Frobenius norm.
\end{theorem}

The minimizer Equation~\eqref{eq:Ustar} directly follows from Equation~\eqref{eq:envelope-isometry}. 
Writing $\widehat{U}=S\odot \widehat{E}$ with $\mathrm{rank}(\widehat{E})\le l$ yields $\|U-\widehat{U}\|_F=\|E_S(U)-\widehat{E}\|_F$, thereby identifying the optimal choice as the best rank-$l$ approximation $\widehat{E}=\mathrm{TSVD}_l(E_S(U))$ according to the Eckart--Young--Mirsky theorem; remodulating with $S$ results in Equation~\eqref{eq:Ustar}.

Theorem~\ref{thm:rankl-error-main} clarifies the role of $l$: for a \emph{fixed} binary basis $S$, MDBF approximates the demodulated envelope $E_S(U)$ using a rank-$l$ matrix and then reattaches the same mask.
Thus, MDBF does not allocate additional degrees of freedom to sign patterns; rather, it enhances capacity by permitting multiple envelope modes in the demodulated domain.
DBF represents the special case $l=1$, which constrains the envelope to rank-one and requires all $R$ columns to share a single envelope direction up to scalar weights.
For $l>1$, the feasible set strictly expands, $\mac{F}_1(S)\subseteq \mac{F}_l(S)$, and the optimal error in Equation \eqref{eq:rankl-error} can only decrease.

\subsection{Layer-Wise Optimization}
\label{sec:opt}
We adopt a \emph{layer-wise} PTQ setting, in which we quantize each weight matrix $W\in\mab{R}^{N\times M}$ independently under strict memory and computational constraints. 
Even with the binary sign bases fixed, optimizing MDBF remains non-convex; the objective is bilinear in the factor matrices, and we impose low-rank constraints on their \emph{demodulated envelopes}. We propose a two-stage pipeline: (i) a closed-form initialization based on Multi-Envelope SVID (MSVID) and (ii) local refinement using alternating updates of ADMM. The resulting factors also provide a robust warmup start for QAT.

\subsubsection{Initialization via Multi-Envelope SVID}
\label{sec:opt-init}
We begin by defining the operator used in both initialization and refinement.
For any matrix $Z \in \mab{R}^{N \times M}$ and any $l$ with $1\le l\le \min(N,M)$, define
\begin{equation}
\label{eq:msvid-def}
\mathrm{MSVID}_{l}[Z] \coloneqq \mathrm{sign}(Z) \odot \mathrm{TSVD}_{l}(|Z|).
\end{equation}
Equivalently, let $S_{Z} \coloneqq \mathrm{sign}(Z)$, and noting that the demodulated envelope satisfies $E_{S_{Z}}(Z)=S_{Z}\odot Z =|Z|$, the operator $\mathrm{MSVID}_l[Z]$ preserves the sign mask $S_Z$ while replacing the envelope with its optimal rank-$l$ approximation. 
The special case $l=1$ corresponds to a rank-one envelope.

Given $W\in\mab{R}^{N\times M}$ and an inner dimension $R$ with $1\le R\le \min(N,M)$, we first compute the rank-$R$ truncated SVD, $W_R = U_R\Sigma_R V_R^{\top}$,
to form balanced factors
\begin{equation}
    \label{eq:mesvid-balanced}
    U_{0} \coloneqq U_R\Sigma_{R}^{\nicefrac{1}{2}} \in\mab{R}^{N\times R},~
    V_{0} \coloneqq V_R\Sigma_R^{\nicefrac{1}{2}}\in\mab{R}^{M\times R},
\end{equation}
so that $W_R=U_0 V_0^{\top}$.
We then initialize MDBF by applying Equation \eqref{eq:msvid-def} to each continuous factor:
\begin{equation}
    \label{eq:mesvid-init-compact}
    \widehat{U}_l^{(0)} \coloneqq \mathrm{MSVID}_l[U_0],~
    \widehat{V}_l^{(0)} \coloneqq \mathrm{MSVID}_l[V_0].
\end{equation}
By construction, $\widehat{U}_{l}^{(0)}$ and $\widehat{V}_{l}^{(0)}$ possess rank-$l$ demodulated envelopes under their induced sign masks, conforming to the MDBF factor structure.

\subsubsection{ADMM Refinement}
\label{sec:opt-admm}
Starting from $\widehat{W}^{(0)}=\widehat{U}_{l}^{(0)}\widehat{V}_{l}^{(0) \top}$, we refine the factors through alternating updates that minimize the reconstruction objective $\|W-UV^{\top}\|_{F}^{2}$ while repeatedly imposing the multi-envelope structure via $\mathrm{MSVID}_l[\cdot]$. Specifically, with $V$ fixed, we conduct a limited number of ADMM inner iterations using an auxiliary variable $\widetilde{U}$ and a scaled dual variable $\Lambda$:
\begin{align}
    \widetilde{U}^{(k+1)}
    &= \ab(WV^{(k)}+\rho (U^{(k)}-\Lambda^{(k)})) \\
    &\times \ab(V^{(k) \top} V^{(k)}+\rho I)^{-1},
    \label{eq:admm-U-tilde}\\
    U^{(k+1)}
    &=
    \mathrm{MSVID}_{l} \ab[\widetilde{U}^{(k+1)}+\Lambda^{(k)}],
    \label{eq:admm-U-prox}\\
    \Lambda^{(k+1)}
    &=
    \Lambda^{(k)}+\widetilde{U}^{(k+1)}-U^{(k+1)},
    \label{eq:admm-U-dual}
\end{align}
where $\rho>0$ and $I$ are the $R \times R$ identity matrix. We update $V$ analogously by applying the same steps to the transposed problem while keeping $U$ fixed.

Since $\mathrm{MSVID}_l[\cdot]$ updates the sign mask via $\mathrm{sign}(\cdot)$, it is not an exact Frobenius projection onto any fixed-mask constraint set, and standard ADMM convergence guarantees do not apply directly. Therefore, we consider the method as an ADMM-inspired scheme that alternates between closed-form least-squares updates and a mask-adaptive rank-$l$ envelope step. Crucially, Theorem \ref{thm:rankl-error-main} provides the theoretical backbone: with strong initialization, subsequent updates function as a refinement mechanism that improves the solution in practice.

\section{Experiments}
\label{sec:experiments}

\begin{table*}[tb]
  \centering
  \caption{Perplexities ($\downarrow$) on WikiText2 for each model. Rows indicate the target bitwidth, and columns correspond to bit-parameters $(l,P)$. DBF uses the same parameter format as $(l,P)=(1,1)$, and LittleBit uses the same format as $(l,P)=(1,2)$.}
  \label{tab:ppl_wikitext2_bits_by_lP}
  \setlength{\tabcolsep}{4pt}
  \begin{tabular}{c|c|cc|ccc}
    \toprule
    \textbf{Model} & \textbf{Bits} &
    $(l{=}1,P{=}1)$ & $(l{=}1,P{=}2)$ &
    $(l{=}2,P{=}1)$ & $(l{=}8,P{=}1)$ & $(l{=}16,P{=}1)$ \\
    \midrule

    \multirow{3}{*}{Tiny LLaMA (1.1B)}
      & 1.00 & 78.14 & 114.84 & \qcol 70.17 & \qcol 70.04 & \qcol \textbf{60.40} \\
      & 1.25 & 21.96 & 37.71  & \qcol 22.50 & \qcol 21.67 & \qcol \textbf{19.98} \\
      & 1.50 & 15.85 & 19.85  & \qcol 16.41 & \qcol 14.94 & \qcol \textbf{14.27} \\
    \midrule

    \multirow{3}{*}{Qwen3 (0.6B)}
      & 1.00 & 343.80 & 773.01 & \qcol 415.91 & \qcol \textbf{262.56} & \qcol 277.12 \\
      & 1.25 & 120.89 & 221.01 & \qcol 121.23 & \qcol 128.35 & \qcol \textbf{111.70} \\
      & 1.50 & 100.84 & 121.72 & \qcol 78.77  & \qcol 85.15  & \qcol \textbf{70.97} \\
    \midrule

    \multirow{3}{*}{LLaMA3.2 (1B)}
      & 1.00 & 118.69 & 131.20 & \qcol 118.17 & \qcol 119.82 & \qcol \textbf{104.28} \\
      & 1.25 & 47.99  & 84.12  & \qcol 42.35  & \qcol \textbf{39.88}  & \qcol 41.46 \\
      & 1.50 & 47.78  & 44.90  & \qcol 39.50  & \qcol 41.06  & \qcol \textbf{38.54} \\
    \midrule

    \multirow{3}{*}{LLaMA2 (7B)}
      & 1.00 & \textbf{27.96} & 37.92 & \qcol 28.01 & \qcol 29.72 & \qcol 28.31 \\
      & 1.25 & 13.40 & 21.67 & \qcol \textbf{12.82} & \qcol 15.83 & \qcol 18.26 \\
      & 1.50 &\textbf{9.81}  & 12.14 & \qcol 9.83  & \qcol 9.94  & \qcol 10.50 \\
    \midrule

    \multirow{3}{*}{LLaMA3 (8B)}
      & 1.00 & 1222.32 & 9118.75 & \qcol \textbf{1024.80} & \qcol 2094.86 & \qcol 1639.64 \\
      & 1.25 & 58.61   & 165.43  & \qcol \textbf{50.13}   & \qcol 92.02   & \qcol 164.24  \\
      & 1.50 & 35.62   & 41.40   & \qcol 36.11   & \qcol \textbf{30.14}   & \qcol 31.70   \\
    \midrule

    \multirow{3}{*}{Qwen3 (8B)}
      & 1.00 & 189.02 & 309.30 & \qcol \textbf{108.90} & \qcol 134.00 & \qcol 115.41 \\
      & 1.25 & \textbf{41.10}  & 101.54 & \qcol 53.24  & \qcol 63.73  & \qcol 69.18  \\
      & 1.50 & 31.72  & 39.02  & \qcol 40.22  & \qcol \textbf{30.09}  & \qcol 32.28  \\
    \bottomrule
  \end{tabular}
\end{table*}

\subsection{Setting}
\label{subsec:setting}

\paragraph{Baselines and Quantization Methods.}
We use representative low-rank binary-factor formats as baselines: DBF, corresponding to $(l,P)=(1,1)$, and LittleBit, corresponding to $(l,P)=(1,2)$.
For each method, we optimize the binary-factor parameters using ADMM for 1,000 outer iterations, with 3 inner ADMM updates per outer step. 
For formats utilizing $P$ decomposed terms, we apply this ADMM procedure independently for each term, resulting in a total of $1{,}000 \times P$ outer iterations.
Following ADMM, we perform an additional gradient-based refinement using Adam for 1,500 steps, with a learning rate of $0.01$. During this refinement stage, we update only the real-valued parameters by minimizing the squared reconstruction loss $\|W-\widehat{W}\|_F^2$.
For each target bit budget of $\{1.00, 1.25, 1.50\}$ bits per weight, we choose the largest inner rank that meets the budget, following \citet{boza2025dbf}.

To enhance accuracy in the low-bit regime, we incorporate an error-compensation technique. Specifically, we adopt quantization error propagation (QEP) \citep{arai2025quantization} with $\alpha=0.5$ and use 512 samples from WikiText2 to estimate the required statistics. Finally, in line with standard practice, we retain the first four and last four layers in full precision to ensure a fair and stable comparison across methods.

\paragraph{Models and Datasets.}
We evaluate the proposed low-rank binary-factor format and baseline variants on widely used open-weight LLM families, including LLaMA2 \citep{touvron2023llama}, LLaMA3 \citep{grattafiori2024llama}, and Qwen.
Specifically, we analyze the standard benchmarks of Tiny LLaMA 1.1B, LLaMA2 7B, LLaMA3 8B, and LLaMA3.2 1B in prior quantization work \citep{boza2025dbf}.
To assess robustness across architectural variations, we also evaluate Qwen3 0.6B and Qwen3 8B, which differ from LLaMA models by incorporating RMSNorm variants in their attention components.

\paragraph{Evaluation.}
We follow established evaluation practices for LLM quantization \citep{boza2025dbf, lee2025littlebit}.
We report perplexity (PPL) on WikiText2 and evaluate downstream zero-shot performance on ARC-Easy, and PIQA.
In the main text, we present the average accuracy across tasks and include per-task results in Appendix~\ref{sec:appendix_experiments}.
All experiments were conducted on a single NVIDIA B200 GPU.

\subsection{Result}

\begin{table*}[t]
  \centering
  \caption{Average zero-shot accuracy ($\uparrow$) on ARC-Easy and PIQA. Rows indicate the target bitwidth, and columns correspond to bit-parameters $(l,P)$. Notably, DBF uses the same parameter format as $(l,P)=(1,1)$, and LittleBit uses the same format as $(l,P)=(1,2)$.}
  \label{tab:avg_acc_bits_by_lP}
  \setlength{\tabcolsep}{4pt}
  \begin{tabular}{c|c|cc|ccc}
    \toprule
    \textbf{Model} & \textbf{Bits} &
    $(l{=}1,P{=}1)$ & $(l{=}1,P{=}2)$ &
    $(l{=}2,P{=}1)$ & $(l{=}8,P{=}1)$ & $(l{=}16,P{=}1)$ \\
    \midrule

    \multirow{3}{*}{Tiny LLaMA (1.1B)}
      & 1.00 & 0.4398 & 0.4355 & \qcol 0.4423 & \qcol \textbf{0.4557} & \qcol 0.4449 \\
      & 1.25 & 0.4945 & 0.4608 & \qcol 0.5049 & \qcol 0.5045 & \qcol \textbf{0.5179} \\
      & 1.50 & 0.5252 & 0.5053 & \qcol 0.5273 & \qcol \textbf{0.5311} & \qcol 0.5259 \\
    \midrule

    \multirow{3}{*}{Qwen3 (0.6B)}
      & 1.00 & 0.4237 & 0.4110 & \qcol \textbf{0.4244} & \qcol 0.4120 & \qcol 0.4228 \\
      & 1.25 & 0.4390 & 0.4213 & \qcol 0.4449 & \qcol 0.4449 & \qcol \textbf{0.4457} \\
      & 1.50 & 0.4519 & 0.4317 & \qcol \textbf{0.4604} & \qcol 0.4544 & \qcol 0.4609 \\
    \midrule

    \multirow{3}{*}{LLaMA3.2 (1B)}
      & 1.00 & 0.4639 & \textbf{0.4716} & \qcol 0.4691 & \qcol 0.4680 & \qcol 0.4691 \\
      & 1.25 & 0.4977 & 0.4842 & \qcol 0.5039 & \qcol \textbf{0.5196} & \qcol 0.5099 \\
      & 1.50 & 0.5082 & 0.5202 & \qcol 0.5222 & \qcol \textbf{0.5386} & \qcol 0.5240 \\
    \midrule

    \multirow{3}{*}{LLaMA2 (7B)}
      & 1.00 & \textbf{0.5263} & 0.4919 & \qcol 0.5217 & \qcol 0.5229 & \qcol 0.5229 \\
      & 1.25 & 0.5981 & 0.5522 & \qcol \textbf{0.6096} & \qcol 0.5767 & \qcol 0.5785 \\
      & 1.50 & 0.6398 & 0.6110 & \qcol \textbf{0.6480} & \qcol 0.6243 & \qcol 0.6341 \\
    \midrule

    \multirow{3}{*}{LLaMA3 (8B)}
      & 1.00 & 0.4078 & 0.3997 & \qcol 0.4104 & \qcol \textbf{0.4185} & \qcol 0.4095 \\
      & 1.25 & 0.4796 & 0.4370 & \qcol \textbf{0.5004} & \qcol 0.4716 & \qcol 0.4443 \\
      & 1.50 & \textbf{0.5208} & 0.5056 & \qcol 0.5059 & \qcol 0.5069 & \qcol 0.5000 \\
    \midrule

    \multirow{3}{*}{Qwen3 (8B)}
      & 1.00 & \textbf{0.4635} & 0.4508 & \qcol 0.4588 & \qcol 0.4486 & \qcol 0.4589 \\
      & 1.25 & 0.4971 & 0.4646 & \qcol \textbf{0.4993} & \qcol 0.4819 & \qcol 0.4854 \\
      & 1.50 & 0.5186 & 0.4880 & \qcol 0.5159 & \qcol \textbf{0.5263} & \qcol 0.5119 \\
    \bottomrule
  \end{tabular}
\end{table*}

\paragraph{Reconstruction Error.}

A central motivation for MDBF is that, under an extremely tight bit budget, additional \emph{sign} diversity may inefficiently utilize capacity compared to enhancing \emph{magnitude} expressivity.
We investigate this by analyzing how layer-wise reconstruction error depends on (i) the envelope rank $l$, which increases magnitude capacity while keeping the shared binary sign bases, and (ii) the residual path $P$, which introduces extra factor terms and additional sign bases, as seen in residual binary decompositions \citep{boza2025dbf, lee2025littlebit}.
We evaluate LLaMA2-7B on three representative Transformer blocks (layers $0$, $15$, and $30$) and compute the reconstruction error for the attention and MLP projection matrices. For each weight matrix $W$ and its quantized approximation $\widehat{W}$, we report the relative Frobenius error $\nicefrac{\|W-\widehat{W}\|_F}{\|W\|_F}$ while sweeping $l \in \{1,2,3,8,16\}$ and $P \in \{1,2,3\}$ at matched bits per weight.
Although larger envelope ranks, e.g., $l \ge 16$, may further improve performance, we defer these settings to future work due to their higher computational cost; notably, $l=16$ is sufficient to demonstrate their effectiveness.

Figure~\ref{fig:relerr_lP_llama7b} shows a consistent trend across layers and modules. 
For any fixed $P$, increasing $l$ reduces reconstruction error, whereas increasing $P$ is generally ineffective within the same bit budget. This behavior supports the design choice of MDBF: to reuse a shared binary carrier and allocate the limited capacity to a richer magnitude structure, rather than duplicating sign bases. In contrast, a larger $P$ necessitates the storage of additional sign matrices, diverting bits from magnitude modeling and potentially increasing approximation error. The effect is strongest in the earliest block, where the gap across $(l,P)$ is most significant among several projections. Since quantization noise introduced early can propagate through the network, minimizing errors in the early layers is particularly important.

\paragraph{Perplexity and Zero-shot Accuracy.}
Table~\ref{tab:ppl_wikitext2_bits_by_lP} demonstrates that MDBF is the strongest across models when comparing matched bits-per-weight, resulting in consistent reductions in perplexity. 
Notably, the improvements are most significant in the extremely low bit regime. Table~\ref{tab:avg_acc_bits_by_lP} also presents downstream zero-shot performance and displays trends closely aligned with those in Table~\ref{tab:ppl_wikitext2_bits_by_lP}. 

Across a fixed bit budget, we find that increasing the envelope rank $l$ contributes more to performance than increasing $P$. Specifically, larger $l$ yields improvements in both perplexity and zero-shot accuracy, while larger $P$ provides more modest enhancements. 
In this regime, our method significantly outperforms the baselines, often by a substantial margin. 
By contrast, for larger models around the 7B scale, the gains begin to saturate. We observe that the quantization error continues to decline, suggesting that the method may be overfitting the calibration data instead of translating these improvements into downstream gains. Exploring whether these trends persist with significantly larger calibration sets is an important direction for future work.

\section{Related Work}
\label{sec:related_work}

\paragraph{Post-Training Quantization at $\sim$4 Bits.}
Most practical low-bit deployments of LLMs rely on PTQ in the roughly 4-bit range. Layer-wise calibration methods, such as GPTQ \citep{li2025gptaq, frantar2022gptq} and AWQ \citep{lin2024awq}, achieve strong accuracy without full retraining by minimizing layer-wise reconstruction errors or by identifying and preserving salient weight channels during calibration. However, when the bit budget increases to $\leq 2$ bits per weight, these techniques often result in significant accuracy degradation, underscoring the need for more structured parameterizations.

\paragraph{Structured Representations for 2--3 Bits.}
To achieve the 2–3 bit regime, recent studies often go beyond purely elementwise quantization to adopt structured weight representations.
A representative line combines incoherence processing with lattice- and trellis-coded quantizers, as seen in QuIP\# \citep{tseng2024quip}, QTIP \citep{tseng2024qtip}, and AQLM \citep{egiazarian2024aqlm}.
By encoding weights using structured codebooks instead of independent rounding, these methods enhance effective expressivity within tight bit budgets.
In this regime, heuristic STE-based adjustments tend to be unstable \citep{long2021learning, ichikawa2025high, yin2019understanding}; therefore, refinement is typically framed as a discrete optimization problem \citep{malinovskii2024pv}.

\paragraph{Methods Targeting 1~Bit and Below.}
At or below 1 bit precision, maintaining model quality generally requires more flexibility than naive rounding permits.
Representative approaches include SVID-based parameterizations \citep{xu2024onebit}, token-adaptive mixtures of scaling factors \citep{jo2024binarymos}, and structured sparsity \citep{dong2025stbllm}.
In parallel, \emph{binary-factor} formats maintain efficient inference primitives \citep{boza2025dbf, lee2025littlebit} by decomposing each weight matrix into two bit-packed sign matrices with low-cost diagonal scaling, resulting in inference primarily governed by 1-bit kernels.
A common theme is the explicit addition of degrees of freedom, such as adaptive scales, learnable sparsity patterns, or residual corrections, to mitigate the information bottleneck caused by extreme quantization.

\section{Conclusion}\label{sec:conclusion}
We study extreme low-bit quantization using binary-factor representations and identify a key limitation of DBF: after sign demodulation, all weight magnitudes are constrained to a single shared envelope. This collapse significantly constrains expressivity and leads to diminishing returns as the inner rank increases, explaining the observed performance saturation. 
To address this bottleneck, we introduce MDBF, which enables deployment-friendly binary computation by preserving shared 1-bit sign bases while replacing the single-envelope with a rank-$l$ demodulated envelope that can represent multiple magnitude modes. We present a practical layer-wise PTQ pipeline featuring closed-form initialization followed by an alternating refinement procedure. 
Experiments across the LLaMA and Qwen model families demonstrate that MDBF improves perplexity and zero-shot accuracy compared to prior binary formats with the same bits per weight, highlighting that limited capacity is more effectively allocated to modeling magnitudes rather than replicating sign patterns.

\section{Limitations}\label{sec:limitation}
We focus on layer-wise PTQ. Accordingly, we do not consider submodule-level~\citep{ichikawa2025lpcd} or block-wise~\citep{tseng2024quip, lee2025littlebit} quantization, end-to-end QAT, or distillation-based methods such as LittleBit~\citep{lee2025littlebit}. 
While these alternatives may yield additional accuracy gains, they typically require significantly greater computational resources and introduce nontrivial engineering overhead.
Our approach minimizes a non-convex objective through heuristic alternating updates. As a result, we do not provide guarantees of global optimality; the best-performing hyperparameter settings, e.g., envelope rank $l$ and decomposition depth $P$, may depend on the optimizer and initialization.
Developing more principled optimization strategies or employing robust discrete general optimization methods \citep{ichikawa2025continuous, ichikawa2025optimization, sun2023revisiting, ichikawa2024controlling} and model-selection criteria for extremely low-bit binary-factor formats remains an important open problem.

\section*{Acknowledgments}
This work was supported by JST BOOST, Japan (Grant No. JPMJBY24D0), and RIKEN Center for AIP. Additionally, this work received support from the Strategic Innovation Promotion Program (SIP)
of the Cabinet Office, Government of Japan.

\bibliography{ref}

@article{touvron2023llama,
  title={Llama: Open and efficient foundation language models},
  author={Touvron, Hugo and Lavril, Thibaut and Izacard, Gautier and Martinet, Xavier and Lachaux, Marie-Anne and Lacroix, Timoth{\'e}e and Rozi{\`e}re, Baptiste and Goyal, Naman and Hambro, Eric and Azhar, Faisal and others},
  journal={arXiv preprint arXiv:2302.13971},
  year={2023}
}

@article{grattafiori2024llama,
  title={The llama 3 herd of models},
  author={Grattafiori, Aaron and Dubey, Abhimanyu and Jauhri, Abhinav and Pandey, Abhinav and Kadian, Abhishek and Al-Dahle, Ahmad and Letman, Aiesha and Mathur, Akhil and Schelten, Alan and Vaughan, Alex and others},
  journal={arXiv preprint arXiv:2407.21783},
  year={2024}
}

@article{xu2024onebit,
  title={Onebit: Towards extremely low-bit large language models},
  author={Xu, Yuzhuang and Han, Xu and Yang, Zonghan and Wang, Shuo and Zhu, Qingfu and Liu, Zhiyuan and Liu, Weidong and Che, Wanxiang},
  journal={arXiv preprint arXiv:2402.11295},
  year={2024}
}

@article{frantar2022gptq,
  title={Gptq: Accurate post-training quantization for generative pre-trained transformers},
  author={Frantar, Elias and Ashkboos, Saleh and Hoefler, Torsten and Alistarh, Dan},
  journal={arXiv preprint arXiv:2210.17323},
  year={2022}
}

@article{tseng2024quip,
  title={Quip\#: Even better llm quantization with hadamard incoherence and lattice codebooks},
  author={Tseng, Albert and Chee, Jerry and Sun, Qingyao and Kuleshov, Volodymyr and De Sa, Christopher},
  journal={arXiv preprint arXiv:2402.04396},
  year={2024}
}

@article{lin2024awq,
  title={Awq: Activation-aware weight quantization for on-device llm compression and acceleration},
  author={Lin, Ji and Tang, Jiaming and Tang, Haotian and Yang, Shang and Chen, Wei-Ming and Wang, Wei-Chen and Xiao, Guangxuan and Dang, Xingyu and Gan, Chuang and Han, Song},
  journal={Proceedings of Machine Learning and Systems},
  volume={6},
  pages={87--100},
  year={2024}
}

@article{chee2023quip,
  title={Quip: 2-bit quantization of large language models with guarantees},
  author={Chee, Jerry and Cai, Yaohui and Kuleshov, Volodymyr and De Sa, Christopher M},
  journal={Advances in Neural Information Processing Systems},
  volume={36},
  pages={4396--4429},
  year={2023}
}

@article{malinovskii2024pv,
  title={Pv-tuning: Beyond straight-through estimation for extreme llm compression},
  author={Malinovskii, Vladimir and Mazur, Denis and Ilin, Ivan and Kuznedelev, Denis and Burlachenko, Konstantin and Yi, Kai and Alistarh, Dan and Richtarik, Peter},
  journal={Advances in Neural Information Processing Systems},
  volume={37},
  pages={5074--5121},
  year={2024}
}

@inproceedings{
arai2025quantization,
title={Quantization Error Propagation: Revisiting Layer-Wise Post-Training Quantization},
author={Yamato Arai and Yuma Ichikawa},
booktitle={The Thirty-ninth Annual Conference on Neural Information Processing Systems},
year={2025},
url={https://openreview.net/forum?id=a3l3K9khbL}
}

@inproceedings{
li2025gptaq,
title={{GPTAQ}: Efficient Finetuning-Free Quantization for Asymmetric Calibration},
author={Yuhang Li and Ruokai Yin and Donghyun Lee and Shiting Xiao and Priyadarshini Panda},
booktitle={Forty-second International Conference on Machine Learning},
year={2025},
url={https://openreview.net/forum?id=QdELyl0FST}
}

@inproceedings{tseng2024qtip,
  title     = {QTIP: Quantization with Trellises and Incoherence Processing},
  author    = {Tseng, Albert and Sun, Qingyao and Hou, David and De Sa, Christopher},
  booktitle = {Advances in Neural Information Processing Systems (NeurIPS)},
  year      = {2024},
  url       = {https://openreview.net/forum?id=7sdkLVuYCU}
}

@inproceedings{egiazarian2024aqlm,
  title     = {Extreme Compression of Large Language Models via Additive Quantization},
  author    = {Egiazarian, Vage and Panferov, Andrei and Kuznedelev, Denis and Frantar, Elias and Babenko, Artem and Alistarh, Dan},
  booktitle = {Proceedings of the 41st International Conference on Machine Learning (ICML)},
  year      = {2024},
  url       = {https://arxiv.org/abs/2401.06118}
}

@inproceedings{malinovskii2024pvtuning,
  title     = {PV-Tuning: Beyond Straight-Through Estimation for Extreme LLM Compression},
  author    = {Malinovskii, Vladimir and Mazur, Denis and Ilin, Ivan and Kuznedelev, Denis and Burlachenko, Konstantin and Yi, Kai and Alistarh, Dan and Richtarik, Peter},
  booktitle = {Advances in Neural Information Processing Systems (NeurIPS)},
  year      = {2024},
  url       = {https://arxiv.org/abs/2405.14852}
}

@inproceedings{jo2024binarymos,
  title     = {Mixture of Scales: Memory-Efficient Token-Adaptive Binarization for Large Language Models},
  author    = {Jo, Dongwon and Kim, Taesu and Kim, Yulhwa and Kim, Jae-Joon},
  booktitle = {Advances in Neural Information Processing Systems (NeurIPS)},
  year      = {2024},
  url       = {https://arxiv.org/abs/2406.12311}
}

@inproceedings{dong2025stbllm,
  title     = {STBLLM: Breaking the 1-Bit Barrier with Structured Binary LLMs},
  author    = {Dong, Peijie and Li, Lujun and Zhong, Yuedong and Du, Dayou and Fan, Ruibo and Chen, Yuhan and Tang, Zhenheng and Wang, Qiang and Xue, Wei and Guo, Yike and Chu, Xiaowen},
  booktitle = {International Conference on Learning Representations (ICLR)},
  year      = {2025},
  url       = {https://openreview.net/forum?id=6XUSDvBFkV}
}

@inproceedings{lee2025littlebit,
  title     = {LittleBit: Ultra Low-Bit Quantization via Latent Factorization},
  author    = {Lee, Banseok and Kim, Dongkyu and You, Youngcheon and Kim, Youngmin},
  booktitle = {Advances in Neural Information Processing Systems (NeurIPS)},
  year      = {2025},
  url       = {https://openreview.net/forum?id=zJzu9evD5K}
}

@article{boza2025dbf,
  title     = {Addition is almost all you need: Compressing neural networks with double binary factorization},
  author    = {Bo\v{z}a, Vladim{\'i}r and Macko, Vladim{\'i}r},
  journal   = {CoRR},
  volume    = {abs/2505.11076},
  year      = {2025},
  url       = {https://arxiv.org/abs/2505.11076},
  doi       = {10.48550/arXiv.2505.11076}
}

@article{ichikawa2024controlling,
  title={Controlling continuous relaxation for combinatorial optimization},
  author={Ichikawa, Yuma},
  journal={Advances in Neural Information Processing Systems},
  volume={37},
  pages={47189--47216},
  year={2024}
}

@inproceedings{sun2023revisiting,
  title={Revisiting sampling for combinatorial optimization},
  author={Sun, Haoran and Goshvadi, Katayoon and Nova, Azade and Schuurmans, Dale and Dai, Hanjun},
  booktitle={International Conference on Machine Learning},
  pages={32859--32874},
  year={2023},
  organization={PMLR}
}

@inproceedings{
ichikawa2025optimization,
title={Optimization by Parallel Quasi-Quantum Annealing with Gradient-Based Sampling},
author={Yuma Ichikawa and Yamato Arai},
booktitle={The Thirteenth International Conference on Learning Representations},
year={2025},
url={https://openreview.net/forum?id=9EfBeXaXf0}
}

@article{
ichikawa2025continuous,
title={Continuous Parallel Relaxation for Finding Diverse Solutions in Combinatorial Optimization Problems},
author={Yuma Ichikawa and Hiroaki Iwashita},
journal={Transactions on Machine Learning Research},
issn={2835-8856},
year={2025},
url={https://openreview.net/forum?id=ix33zd5zCw},
note={}
}

@article{ichikawa2025high,
  title={High-Dimensional Learning Dynamics of Quantized Models with Straight-Through Estimator},
  author={Ichikawa, Yuma and Kashiwamura, Shuhei and Sakata, Ayaka},
  journal={arXiv preprint arXiv:2510.10693},
  year={2025}
}

@article{yin2019understanding,
  title={Understanding straight-through estimator in training activation quantized neural nets},
  author={Yin, Penghang and Lyu, Jiancheng and Zhang, Shuai and Osher, Stanley and Qi, Yingyong and Xin, Jack},
  journal={arXiv preprint arXiv:1903.05662},
  year={2019}
}

@article{long2021learning,
  title={Learning quantized neural nets by coarse gradient method for nonlinear classification},
  author={Long, Ziang and Yin, Penghang and Xin, Jack},
  journal={Research in the Mathematical Sciences},
  volume={8},
  number={3},
  pages={48},
  year={2021},
  publisher={Springer}
}

@article{ichikawa2025lpcd,
  title={LPCD: Unified Framework from Layer-Wise to Submodule Quantization},
  author={Ichikawa, Yuma and Fujimoto, Yudai and Sakai, Akira},
  journal={arXiv preprint arXiv:2512.01546},
  year={2025}
}

@inproceedings{roy2007effectiveRank,
  title     = {The Effective Rank: A Measure of Effective Dimensionality},
  author    = {Roy, Olivier and Vetterli, Martin},
  booktitle = {15th European Signal Processing Conference (EUSIPCO 2007)},
  pages     = {606--610},
  year      = {2007},
  address   = {Poznan, Poland},
  month     = sep,
  url       = {https://www.eurasip.org/Proceedings/Eusipco/Eusipco2007/Papers/a5p-h05.pdf},
  note      = {Event date: September 3--7, 2007}
}

\newpage

\appendix
\onecolumn

\section{Derivation}
\label{app-sec:derivation}

\subsection{Proofs of Main Results}
\label{sec:appendix-proof}

This section presents detailed proofs of the results stated in the main text, along with the necessary auxiliary lemmas for completeness.

\subsubsection{Demodulated envelopes and Envelope Class}

\paragraph{Demodulated Envelope.}
Let $S\in\{\pm1\}^{p\times q}$ represent a sign mask and $Z\in\mab{R}^{p\times q}$ an arbitrary matrix. We define the \emph{demodulated envelope} of $Z$ concerning $S$ as follows:
\begin{equation}
    \label{eq:demod-envelope-app}
    E_S(Z)\coloneqq S\odot Z,
\end{equation}
Since $S_{ij}^2=1$ holds for all $(i,j)$, we have $S\odot S=\mathbf{1}$ entrywise; therefore, $Z = S\odot E_S(Z)$.

\paragraph{Envelope Class.}
We recall the rank-$l$ envelope class associated with a fixed sign mask.
\begin{definition}
\label{def:Fl-app}
Let $S\in\{\pm1\}^{p\times q}$ and $l\ge 1$. Define
\begin{equation}
    \label{eq:Fl-rank-form}
    \mac{F}_l(S)\coloneqq
    \left\{
    S\odot E  \mid E\in\mab{R}^{p\times q}, \mathrm{rank}(E)\le l
    \right\}
    \subset \mab{R}^{p\times q}.
\end{equation}
Equivalently, $\widehat{Z}\in\mac{F}_l(S)$ if and only if its demodulated envelope $E_S(\widehat{Z})$ has a rank of at most $l$.
\end{definition}

The next lemma demonstrates that $\mac{F}_l(S)$ can be equivalently characterized as a sum of diagonal scalings of the fixed mask $S$, thereby relating envelopes to DBF-like parameterizations.
\begin{lemma}
\label{lem:diag-form}
Let $S\in\{\pm1\}^{p\times q}$ and $l\ge 1$. Then
\begin{equation}
    \label{eq:Fl-diag-form}
    \mac{F}_l(S)
    =
    \left\{
    \sum_{t=1}^{l} D_{\B{x}^{(t)}} S D_{\B{y}^{(t)}} \mid
    \B{x}^{(t)}\in\mab{R}^{p}, \B{y}^{(t)}\in\mab{R}^{q}
    \right\}.
\end{equation}
\end{lemma}

\begin{proof}
    We prove both inclusions.
    Let $\widehat{Z}\in\mac{F}_l(S)$. By Definition \ref{def:Fl-app}, there exists $E\in\mab{R}^{p\times q}$ such that $\mathrm{rank}(E)\le l$ with $\widehat{Z}=S\odot E$.
    Since $\mathrm{rank}(E)\le l$, there exist vectors $\{\B{x}^{(t)}\}_{t=1}^l\subset\mab{R}^p$ and $\{\B{y}^{(t)}\}_{t=1}^l\subset\mab{R}^q$ such that $
        E=\sum_{t=1}^l \B{x}^{(t)}(\B{y}^{(t)})^\top.$
    For any $(i,j)$, we compute
    \begin{align}
        \widehat{Z}_{ij}
        &=(S\odot E)_{ij}
        =S_{ij}E_{ij}
        =S_{ij}\sum_{t=1}^l x_i^{(t)}y_j^{(t)}
        =\sum_{t=1}^l x_i^{(t)}S_{ij}y_j^{(t)} =\sum_{t=1}^l \bigl(D_{\B{x}^{(t)}} S D_{\B{y}^{(t)}}\bigr)_{ij}.
    \end{align}
    Therefore, $\widehat{Z}=\sum_{t=1}^l D_{\B{x}^{(t)}} S D_{\B{y}^{(t)}}$, which shows that $\widehat{Z}$ belongs to the right-hand side of Equation \eqref{eq:Fl-diag-form}.
    Suppose that $
        \widehat{Z}=\sum_{t=1}^l D_{\B{x}^{(t)}} S D_{\B{y}^{(t)}}$.
    Define $E\coloneqq \sum_{t=1}^l \B{x}^{(t)}\bigl(\B{y}^{(t)}\bigr)^\top$. 
    Then $\mathrm{rank}(E)\le l$. Moreover, for any $(i,j)$,
    \begin{align}
        (S\odot E)_{ij}
        &=S_{ij}E_{ij}
        =S_{ij}\sum_{t=1}^l x_i^{(t)}y_j^{(t)}
        =\sum_{t=1}^l x_i^{(t)}S_{ij}y_j^{(t)}
        =\sum_{t=1}^l \bigl(D_{\B{x}^{(t)}} S D_{\B{y}^{(t)}}\bigr)_{ij}
        =\widehat{Z}_{ij}.
    \end{align}
    Hence, $\widehat{Z}=S\odot E\in\mac{F}_l(S)$ completes the proof.
\end{proof}

\subsubsection{Isometry Induced by Sign Mask}

\begin{lemma}
    \label{lem:sign-invariance}
    Let $S\in\{\pm1\}^{p\times q}$ and $A,B\in\mab{R}^{p\times q}$.
    \begin{equation}
        \|S\odot A\|_{F}=\|A\|_{F},~~
        \|S\odot(A-B)\|_{F}=\|A-B\|_{F}.
    \end{equation}
 The linear map $T_S:\mab{R}^{p\times q}\to \mab{R}^{p\times q}$ defined by $T_S(Z)\coloneqq S\odot Z$
    is an isometry with respect to $\|\cdot\|_{F}$.
\end{lemma}

\begin{proof}
    Using the definition of the Frobenius norm and the fact that $S_{ij}^2=1$ for all $(i,j)$, we have
    \begin{align}
        \|S\odot A\|_{F}^2
        &=\sum_{i=1}^{p}\sum_{j=1}^{q}(S_{ij}A_{ij})^2
        =\sum_{i=1}^{p}\sum_{j=1}^{q}S_{ij}^2\,A_{ij}^2
        =\sum_{i=1}^{p}\sum_{j=1}^{q}A_{ij}^2
        =\|A\|_{F}^{2}.
    \end{align}
    Taking the square roots yields $\|S\odot A\|_F=\|A\|_F$. Applying the same argument to $A$-$B$ yields
    \begin{equation}
        \|S\odot(A-B)\|_F=\|A-B\|_F,
    \end{equation}
    which establishes the distance-preservation property and, consequently, the isometry claim.
\end{proof}

\subsubsection{Optimal Demodulated-Envelope Approximation}\label{subsubsec:optimal-demodulated}

We restate and prove Theorem~\ref{thm:rankl-error-main} from the perspective of the demodulated-envelope.

\begin{theorem}
    \label{thm:rankl-error-app}
    Let $U\in \mab{R}^{N\times R}$ fix a sign mask $S\in\{\pm1\}^{N\times R}$.
    Let $\sigma_1(E_S(U))\ge\cdots\ge\sigma_{\min(N,R)}(E_S(U))\ge 0$
    denote the singular values of the demodulated envelope $E_S(U)\coloneqq S\odot U$.
    Fix an integer $l$ with $1\le l\le \min(N,R)$.
    Then
    \begin{equation}
    \label{eq:rankl-error-app}
    \min_{\widehat{U}\in\mac{F}_l(S)} \|U-\widehat{U}\|_{F}^2
    =
    \sum_{i=l+1}^{\min(N, R)}\sigma_i\bigl(E_S(U)\bigr)^2.
    \end{equation}
    Moreover, a minimizer is provided by
    \begin{equation}
    \label{eq:rankl-opt-app}
    \widehat{U}^{\star}=S\odot \mathrm{TSVD}_l\bigl(E_S(U)\bigr),
    \end{equation}
    where $\mathrm{TSVD}_l(\cdot)$ denotes the best rank-$l$ approximation in the Frobenius norm.
\end{theorem}

\begin{proof}
    By Definition~\ref{def:Fl}, $\widehat{U}\in\mac{F}_l(S)$ if and only if there exists a matrix
    $E\in\mab{R}^{N\times R}$ such that $\mathrm{rank}(E)\le l$ as follows:
    \begin{equation}
        \label{eq:proof-param}
        \widehat{U}=S\odot E.
    \end{equation}
    Hence, the constrained minimization over $\widehat{U}\in\mac{F}_l(S)$ is equivalent to minimizing over
    $E$ with $\mathrm{rank}(E)\le l$ under the parameterization in Equation~\eqref{eq:proof-param}.
    Define $E_U\coloneqq E_S(U)=S\odot U$. Since $S_{ij}^2=1$ for all $(i,j)$, we have the involution property
    $S\odot(S\odot Z)=Z$ for any $Z$, particularly $U=S\odot E_U$.
    For any feasible $\widehat{U}=S\odot E$, we obtain $U-\widehat{U}=S\odot(E_U-E)$.
    Applying Lemma~\ref{lem:sign-invariance} yields
    \begin{equation}
        \|U-\widehat{U}\|_F=\|E_U-E\|_F,~~
        \|U-\widehat{U}\|_F^2=\|E_U-E\|_F^2.
        \label{eq:reduce-rank-prob}
    \end{equation}
    Substituting Equations \eqref{eq:proof-param} and \eqref{eq:reduce-rank-prob} gives
    \begin{equation}
        \min_{\widehat{U}\in\mac{F}_l(S)} \|U-\widehat{U}\|_{F}^2
        =
        \min_{\substack{E\in\mab{R}^{N\times R} \\ \mathrm{rank}(E)\le l}}
        \|E_U-E\|_{F}^2,
    \end{equation}
    which is exactly the classical best rank-$l$ approximation problem for $E_U$ in Frobenius norm.
    By the Eckart–Young–Mirsky theorem applied to $E_U$, we have
    \begin{equation}
        \min_{\mathrm{rank}(E)\le l}\|E_U-E\|_{F}^2 =
        \sum_{i=l+1}^{\min(N,R)}\sigma_i(E_U)^2 =
        \sum_{i=l+1}^{\min(N,R)}\sigma_i\bigl(E_S(U)\bigr)^2,
    \end{equation}
    which proves Equation \eqref{eq:rankl-error-app}. Moreover, the same theorem implies that an optimizer is
    $E^\star=\mathrm{TSVD}_l(E_U)$; substituting into Equation \eqref{eq:proof-param} yields
    $\widehat{U}^\star=S\odot \mathrm{TSVD}_l(E_U)=S\odot \mathrm{TSVD}_l(E_S(U))$, which corresponds to Equation \eqref{eq:rankl-opt-app}.
\end{proof}
\begin{remark}
    If we choose the mask $S=\mathrm{sign}(U)$, the demodulated envelope satisfies
    $E_S(U)=S\odot U=|U|$ entrywise. In this case, Equation \eqref{eq:rankl-error-app} specializes to the standard singular-value tail characterization of the best rank-$l$ approximation error for $|U|$ in the Frobenius norm.
\end{remark}

\begin{corollary}
    \label{cor:dbf-l1-app}
    Under the assumptions of Theorem~\ref{thm:rankl-error-app}, setting $l=1$ yields
    \begin{equation}
        \min_{\widehat{U}\in\mac{F}_1(S)} \|U-\widehat{U}\|_{F}^2
        =
        \sum_{i=2}^{\min(N,M)}\sigma_i\bigl(E_S(U)\bigr)^2,.
    \end{equation}
    In particular, the single envelope constraint is lossless (in Frobenius norm) if and only if
    $\mathrm{rank}\bigl(E_S(U)\bigr)\le 1$.
\end{corollary}

\begin{proof}
    This is the specialization of Theorem~\ref{thm:rankl-error-app} to $l=1$.
\end{proof}

\subsubsection{Exact Projection}

\begin{proposition}
    \label{prop:proj-Fl}
    Fix $S\in\{\pm1\}^{N\times R}$ and $l\ge 1$.
    For any $Z\in\mab{R}^{N\times R}$, the Frobenius projection onto $\mac{F}_l(S)$,
    \begin{equation}
        \Pi_{\mac{F}_l(S)}(Z)\in \arg\min_{Y\in\mac{F}_l(S)}\|Z-Y\|_F,
    \end{equation}
    selected as
    \begin{equation}
        \label{eq:proj-Fl-app}
        \Pi_{\mac{F}_l(S)}(Z) = S\odot \mathrm{TSVD}_l\bigl(E_S(Z)\bigr) = S\odot \mathrm{TSVD}_l\bigl(S\odot Z\bigr).
    \end{equation}
\end{proposition}

\begin{proof}
    By Definition~\ref{def:Fl-app}, any $Y\in\mac{F}_l(S)$ can be expressed as $Y=S\odot E$ for some
    $E\in\mab{R}^{N\times R}$ with $\mathrm{rank}(E)\le l$. Hence,
    \begin{align}
        \min_{Y\in\mac{F}_l(S)}\|Z-Y\|_F^2
        &= \min_{\substack{E\in\mab{R}^{N\times R}\\ \mathrm{rank}(E)\le l}} \|Z-S\odot E\|_F^2.
    \end{align}
    Using the sign-mask isometry (Lemma~\ref{lem:sign-invariance}) and the identity $S\odot(S\odot E)=E$, we obtain
    \begin{align}
        \|Z-S\odot E\|_F
        &=\|S\odot Z - S\odot(S\odot E)\|_F
        =\|S\odot Z - E\|_F
        =\|E_S(Z)-E\|_F.
    \end{align}
    Therefore,
    \begin{equation}
        \min_{Y\in\mac{F}_l(S)}\|Z-Y\|_F^2
        =
        \min_{\mathrm{rank}(E)\le l}\|E_S(Z)-E\|_F^2.
    \end{equation}
    The right-hand side represents the classical best rank-$l$ approximation problem for $E_S(Z)$ in the Frobenius norm, with minimizers provided by the truncated SVD:
    $E^\star=\mathrm{TSVD}_l(E_S(Z))$ (Eckart–Young–Mirsky theorem). Substituting back yields
    $Y^\star=S\odot E^\star=S\odot \mathrm{TSVD}_l (E_S(Z))$, which is Equation \eqref{eq:proj-Fl-app}.
\end{proof}

\section{Additional Implementation Details}
\label{sec:appendix_imp_detail}

\subsection{Entropy-based effective rank}
\label{app:effective_rank}

We quantify the number of envelope modes present using the \emph{entropy-based effective rank} (entropy rank) \citep{roy2007effectiveRank}.
For a matrix $E\in\mab{R}^{N \times M}$ (in our case, a demodulated envelope such as
$E_{\mathrm{sign}(U)}(U)=|U|$ or $E_{\mathrm{sign}(V)}(V)=|V|$), let singular values and $\sigma_1(E)\ge \cdots \ge \sigma_{\min(N,M)}(E)\ge 0$ represent its normalized spectrum.
Define the normalized spectrum
\begin{equation}
    \pi_i(E)\coloneqq \frac{\sigma_i(E)}{\sum_{j=1}^{\min(N,M)}\sigma_j(E)},~~~ i=1,\dots,\min(N,M),
\end{equation}
with the convention $\pi_i(E)=1/\min(N,M)$, if $E=0$ and the spectral entropy are considered:
\begin{equation}
    H(E)\coloneqq -\sum_{i=1}^{\min(N,M)}\pi_i(E)\log \pi_i(E).
\end{equation}
The effective rank of entropy is $\mathrm{erank}(E)\coloneqq \exp\bigl(H(E)\bigr)$.
This quantity is scale-invariant and satisfies $1\le \mathrm{erank}(E)\le \min(N,M)$.
Moreover, $\mathrm{erank}(E)=1$ when $E$ is a nonzero rank-one matrix with all spectral mass on a single singular value, and $\mathrm{erank}(E)\approx k$ when roughly $k$ singular values contribute comparably.

\section{Additional Experiments}
\label{sec:appendix_experiments}

In the main text, we present results aggregated across models (and, where applicable, across tasks) to emphasize overall trends. For completeness and transparency, this appendix includes the \emph{individual} results for each model and target bitwidth. All experiments follow the same evaluation protocol as the main paper; the reported metric remains the same normalized accuracy (higher is better).

Tables~\ref{tab:arc_easy_acc_norm_bits_by_lP} and~\ref{tab:piqa_acc_norm_bits_by_lP} summarize normalized accuracy on ARC-Easy and PIQA, respectively. Rows indicate the target bit width, while columns correspond to the bit-parameter choices $(l,P)$. Overall, the per-model numbers align with the aggregate results in the main text: changing $(l,P)$ typically results in modest variations, with $(l{=}2,P{=}1)$ providing a strong and stable default across models, while increasing $l$ often leads to diminishing returns.

\begin{table*}[t]
  \centering
  \caption{
  Zero-shot accuracy ($\uparrow$) on ARC-Easy. Rows indicate the target bitwidth, and columns correspond to bit-parameters $(l,P)$.
  }
  \label{tab:arc_easy_acc_norm_bits_by_lP}
  \setlength{\tabcolsep}{4pt}
  \begin{tabular}{c|c|cc|ccc}
    \toprule
    \textbf{Model} & \textbf{Bits} &
    $(l{=}1,P{=}1)$ & $(l{=}1,P{=}2)$ &
    $(l{=}2,P{=}1)$ & $(l{=}8,P{=}1)$ & $(l{=}16,P{=}1)$ \\
    \midrule

    \multirow{3}{*}{Tiny LLaMA (1.1B)} & 1.00 & 0.3279 & 0.3258 & \qcol 0.3215 & \qcol \textbf{0.3304} & \qcol 0.3283 \\
     & 1.25 & 0.3737 & 0.3384 & \qcol \textbf{0.3960} & \qcol 0.3822 & \qcol 0.3927 \\
     & 1.50 & 0.4057 & 0.3838 & \qcol 0.4045 & \qcol 0.4066 & \qcol \textbf{0.4082} \\
    \midrule

    \multirow{3}{*}{Qwen3 (0.6B)} & 1.00 & 0.3051 & 0.2883 & \qcol \textbf{0.3157} & \qcol 0.2837 & \qcol 0.3064 \\
     & 1.25 & 0.3258 & 0.3089 & \qcol 0.3283 & \qcol \textbf{0.3338} & \qcol 0.3245 \\
     & 1.50 & 0.3418 & 0.3140 & \qcol 0.3565 & \qcol 0.3401 & \qcol \textbf{0.3485} \\
    \midrule

    \multirow{3}{*}{LLaMA3.2 (1B)} & 1.00 & 0.3413 & \textbf{0.3653} & \qcol 0.3359 & \qcol 0.3523 & \qcol 0.3472 \\
     & 1.25 & 0.3843 & 0.3540 & \qcol 0.3897 & \qcol 0.3956 & \qcol \textbf{0.3990} \\
     & 1.50 & 0.3906 & 0.4217 & \qcol 0.4112 & \qcol \textbf{0.4369} & \qcol 0.4082 \\
    \midrule

    \multirow{3}{*}{LLaMA2 (7B)} & 1.00 & 0.4221 & 0.3842 & \qcol 0.4209 & \qcol 0.4228 & \qcol \textbf{0.4265} \\
     & 1.25 & 0.5118 & 0.4482 & \qcol \textbf{0.5189} & \qcol 0.4853 & \qcol 0.4899 \\
     & 1.50 & 0.5614 & 0.5240 & \qcol \textbf{0.5800} & \qcol 0.5370 & \qcol 0.5631 \\
    \midrule

    \multirow{3}{*}{LLaMA3 (8B)} & 1.00 & 0.2879 & 0.2761 & \qcol 0.3001 & \qcol \textbf{0.3077} & \qcol 0.2946 \\
     & 1.25 & 0.3628 & 0.3266 & \qcol \textbf{0.3817} & \qcol 0.3544 & \qcol 0.3232 \\
     & 1.50 & \textbf{0.4099} & 0.4024 & \qcol 0.3965 & \qcol 0.3864 & \qcol 0.3851 \\
    \midrule

    \multirow{3}{*}{Qwen3 (8B)} & 1.00 & \textbf{0.3514} & 0.3439 & \qcol 0.3443 & \qcol 0.3481 & \qcol \textbf{0.3514} \\
     & 1.25 & \textbf{0.4045} & 0.3645 & \qcol 0.4007 & \qcol 0.3788 & \qcol 0.3805 \\
     & 1.50 & \textbf{0.4196} & 0.3885 & \qcol 0.4415 & \qcol 0.4373 & \qcol 0.4188 \\
    \bottomrule
  \end{tabular}
\end{table*}

\begin{table*}[t]
  \centering
  \caption{Zero-shot accuracy ($\uparrow$) on PIQA. Rows indicate the target bitwidth, and columns correspond to bit-parameters $(l,P)$.}
  \label{tab:piqa_acc_norm_bits_by_lP}
  \setlength{\tabcolsep}{4pt}
  \begin{tabular}{c|c|cc|ccc}
    \toprule
    \textbf{Model} & \textbf{Bits} &
    $(l{=}1,P{=}1)$ & $(l{=}1,P{=}2)$ &
    $(l{=}2,P{=}1)$ & $(l{=}8,P{=}1)$ & $(l{=}16,P{=}1)$ \\
    \midrule

    \multirow{3}{*}{Tiny LLaMA (1.1B)} & 1.00 & 0.5517 & 0.5452 & \qcol 0.5631 & \qcol \textbf{0.5811} & \qcol 0.5615 \\
     & 1.25 & 0.6153 & 0.5832 & \qcol 0.6137 & \qcol 0.6268 & \qcol \textbf{0.6431} \\
     & 1.50 & 0.6447 & 0.6268 & \qcol 0.6502 & \qcol \textbf{0.6556} & \qcol 0.6436 \\
    \midrule

    \multirow{3}{*}{Qwen3 (0.6B)} & 1.00 & \textbf{0.5424} & 0.5337 & \qcol 0.5332 & \qcol 0.5403 & \qcol 0.5392 \\
     & 1.25 & 0.5522 & 0.5337 & \qcol 0.5615 & \qcol 0.5560 & \qcol \textbf{0.5669} \\
     & 1.50 & 0.5620 & 0.5495 & \qcol 0.5642 & \qcol 0.5686 & \qcol \textbf{0.5734} \\
    \midrule

    \multirow{3}{*}{LLaMA3.2 (1B)} & 1.00 & 0.5865 & 0.5778 & \qcol \textbf{0.6023} & \qcol 0.5838 & \qcol 0.5909 \\
     & 1.25 & 0.6110 & 0.6143 & \qcol 0.6181 & \qcol \textbf{0.6436} & \qcol 0.6208 \\
     & 1.50 & 0.6257 & 0.6186 & \qcol 0.6333 & \qcol \textbf{0.6404} & \qcol 0.6398 \\
    \midrule

    \multirow{3}{*}{LLaMA2 (7B)} & 1.00 & \textbf{0.6306} & 0.5997 & \qcol 0.6225 & \qcol 0.6230 & \qcol 0.6192 \\
     & 1.25 & 0.6844 & 0.6561 & \qcol \textbf{0.7002} & \qcol 0.6681 & \qcol 0.6670 \\
     & 1.50 & \textbf{0.7182} & 0.6980 & \qcol 0.7160 & \qcol 0.7116 & \qcol 0.7051 \\
    \midrule

    \multirow{3}{*}{LLaMA3 (8B)} & 1.00 & 0.5277 & 0.5234 & \qcol 0.5207 & \qcol \textbf{0.5294} & \qcol 0.5245 \\
     & 1.25 & 0.5963 & 0.5473 & \qcol \textbf{0.6192} & \qcol 0.5887 & \qcol 0.5653 \\
     & 1.50 & \textbf{0.6317} & 0.6088 & \qcol 0.6153 & \qcol 0.6273 & \qcol 0.6148 \\
    \midrule

    \multirow{3}{*}{Qwen3 (8B)} & 1.00 & \textbf{0.5756} & 0.5577 & \qcol 0.5734 & \qcol 0.5490 & \qcol 0.5664 \\
     & 1.25 & 0.5898 & 0.5647 & \qcol \textbf{0.5979} & \qcol 0.5849 & \qcol 0.5903 \\
     & 1.50 & \textbf{0.6175} & 0.5876 & \qcol 0.5903 & \qcol 0.6153 & \qcol 0.6050 \\
    \bottomrule
  \end{tabular}
\end{table*}

\end{document}